\documentclass[10pt,twocolumn,letterpaper]{article}

\usepackage{cvpr}              % To produce the CAMERA-READY version
\usepackage[accsupp]{axessibility}
\usepackage{graphicx}
\usepackage{amsmath}
\usepackage{amssymb}
\usepackage{booktabs}
\usepackage{algorithm}
\usepackage{listings}
\usepackage{color}
\usepackage{boldline}
\usepackage{ulem}
\usepackage{multirow}
\usepackage{bbm}
\usepackage{bm}
\usepackage{soul}
\definecolor{r}{RGB}{202,12,22} 
\usepackage[pagebackref,breaklinks,colorlinks]{hyperref}

\usepackage[capitalize]{cleveref}
\crefname{section}{Sec.}{Secs.}
\Crefname{section}{Section}{Sections}
\Crefname{table}{Table}{Tables}
\crefname{table}{Tab.}{Tabs.}

\def\cvprPaperID{2811} % *** Enter the CVPR Paper ID here
\def\confName{CVPR}
\def\confYear{2022}

\begin{document}

% \title{\vspace{-1.em} Recurrent Dynamic Embedding for Video Object Segmentation
% \vspace{-.5em}}
\title{Recurrent Dynamic Embedding for Video Object Segmentation}

\author{
Mingxing Li$^{1*}$,
Li Hu$^{2*}$,
Zhiwei Xiong$^{1\dag}$,
Bang Zhang$^{2}$,
Pan Pan$^{2}$,
Dong Liu$^{1}$
\\
$^1$University of Science and Technology of China \\
$^2$Alibaba DAMO Academy, Alibaba Group \\ 
{\tt\small mxli@mail.ustc.edu.cn \{zwxiong, dongeliu\}@ustc.edu.cn}\\
{\tt\small\{hooks.hl, zhangbang.zb, panpan.pp\}@alibaba-inc.com}
}
\maketitle

%%%%%%%%% ABSTRACT
\begin{abstract}
\renewcommand{\thefootnote}{}
\footnotetext{$^*$ Equal contribution.  $^\dag$ Corresponding author. This work was done during Mingxing Li's internship at Alibaba.}
\vspace{-.5em}
Space-time memory (STM) based video object segmentation (VOS) networks usually keep increasing memory bank every several frames, which shows excellent performance. However, 1) the hardware cannot withstand the ever-increasing memory requirements as the video length increases. 2) Storing lots of information inevitably introduces lots of noise, which is not conducive to reading the most important information from the memory bank. In this paper, we propose a Recurrent Dynamic Embedding (RDE) to build a memory bank of constant size. Specifically, we explicitly generate and update RDE by the proposed Spatio-temporal Aggregation Module (SAM), which exploits the cue of historical information. To avoid error accumulation owing to the recurrent usage of SAM, we propose an unbiased guidance loss during the training stage, which makes SAM more robust in long videos. Moreover, the predicted masks in the memory bank are inaccurate due to the inaccurate network inference, which affects the segmentation of the query frame. To address this problem, we design a novel self-correction strategy so that the network can repair the embeddings of masks with different qualities in the memory bank. Extensive experiments show our method achieves the best tradeoff between performance and speed. Code is available 
at \url{https://github.com/Limingxing00/RDE-VOS-CVPR2022}.
% \vspace{-1.5em}
\end{abstract}

%%%%%%%%% BODY TEXT

\section{Introduction\label{sec:intro}}
% \vspace{-0.3em}
Video object segmentation (VOS) is a fundamental task for video understanding, including lots of applications, such as autonomous driving and video editing. This work focuses on semi-supervised VOS setting. In this setting, given the instances annotation of the first frame, the VOS algorithms segment the instances in other frames.

\begin{figure}[t]
\centering % 图片居中
\includegraphics[width=.95\linewidth]{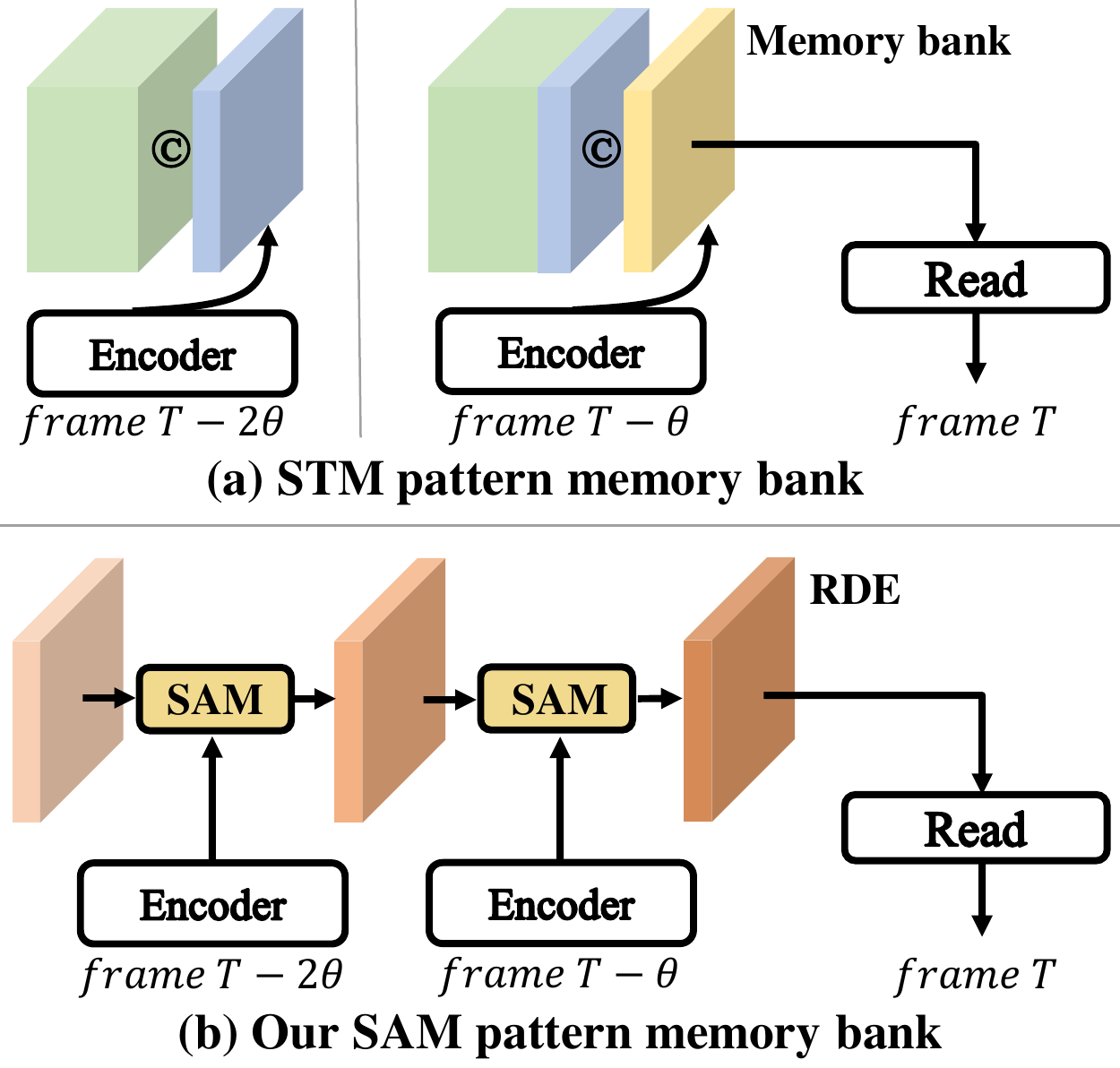}
\vspace{-.3em}
% \caption{An illustration of the memory bank update mechanism. (a) denotes an initial state of the memory bank embedding, which is presented by blue points. (b) shows STM pattern update in the concatenation way, where red points represent new embeddings. (c) denotes exponential moving average (EMA) pattern, where some old embeddings merge with the new compatible ones in the EMA way. In contrast, our SAM (d) aggregates both old and new embeddings in a learning manner.}
\caption{The inference pipelines of the segmentation of frame $T$. \raisebox{.4pt}{\textcircled{\raisebox{-.8pt} {c}}} denotes concatenation. $\theta$ denotes the sampling interval for the update of the memory bank. (a) shows the network read the space-time memory (STM) pattern memory bank to segment frame $T$. As the length of videos increases, the STM pattern memory bank has an ever-increasing size. In (b), we update a recurrent dynamic embedding (RDE) to build a memory bank of the constant size, which is maintained by a spatio-temporal aggregation module (SAM).}
\label{fig:fig1}
\vspace{-1em}
\end{figure}
% \begin{figure}[!t]
% \centering % 图片居中
% \includegraphics[width=1\linewidth]{figure/contrast.pdf}
% \vspace{-.5em}
% \caption{Previous matching based VOS methods (a-c)  for the segmentation of the query frame. Our method (d) utilizes a \textbf{learnable way} to support the \textbf{constant size} memory bank recurrently. EMA denotes exponential moving average, which is an unlearnable operation.}
% \label{fig:contrast}
% \vspace{-1em}
% \end{figure}

Matching based networks \cite{oh2018fast,yang2018efficient,lu2020video, oh2019video, seong2020kernelized, zhang2020transductive, cheng2021modular, hu2021learning,liang2021video,mao2021joint,seong2021hierarchical} are popular for semi-supervised VOS. These networks have a memory bank mechanism, which encodes some frames into embeddings and stores those embeddings in the memory bank to assist the segmentation of the query frame.
% Previous matching based VOS methods adopting the memory bank can be roughly divided into three categories, 1) Several frame dependent methods, which utilize the limited embeddings of 
Some methods only use the embeddings of a limited number of frames, such as 
the ground-truth (GT) frame \cite{hu2018videomatch}, the latest frame
(for brevity, the latest frame of the query frame is abbreviated as the latest frame) \cite{perazzi2017learning} and both of them \cite{oh2018fast,yang2018efficient,lu2020video}. These methods do not make full use of historical frames in the video. STM based methods \cite{oh2019video, seong2020kernelized,zhang2020transductive,cheng2021modular,hu2021learning,liang2021video,mao2021joint,seong2021hierarchical} store the embeddings every several (e.g., 5) frames in the STM pattern memory bank as shown in Figure \ref{fig:fig1}(a). Although STM based methods utilize equal interval sampling to mine the historical information in the video, as the length of videos increases, the STM pattern memory bank has an ever-increasing size and inevitably introduces lots of noise. Exponential moving average (EMA) based methods \cite{li2020fast, liang2020video, wang2021swiftnet} try to address the problems. The EMA based methods index some pixel embeddings from the embeddings of the query frame and the memory bank according to certain criteria and fuse these pixel embeddings in the EMA way.  However, the EMA based methods have a strong limitation because of the direct summation operation (see details in Sec. \ref{subsec:revisit}).

% While the STM based methods have ever-increasing memory requirements and inevitably introduces lots of noise. The EMA based methods try to address these problems  by the fusion of embeddings in the EMA way. However, the EMA based methods can only perform between the most similar embeddings by the index operation due to the direct summation operation of EMA (see details in Sect. \ref{subsec:revisit}), which is an unlearnable operation.

In this paper, we address two problems. 1) How to build and update a memory bank of the constant size to effectively and efficiently store historical information? 2) Except for the GT frame, other masks are inaccurate owing to the inaccurate network inference, how to correct the poor embedding encoded from the inaccurate masks?

% We find most of the methods have a conclusion that only using the embedding of the latest frame is better than only using the embedding of the GT frame to segment the query frame.  We think the latest frame is usually the most similar to the query frame. Empirically, \textit{the segmentation quality of the query frame is affected by the difference between the instances of the query frame and the memory bank}.
% For example, when the memory bank includes a person standing frontally, and the query frame is a person with a similar movement, the segmentation result is usually satisfactory. While when the query frame is a person with a dissimilar movement, the segmentation result is usually poor.  In addition, we think the mask quality of a frame stored in the memory bank also affects the segmentation of the query frame. However, the previous methods have  hardly been explored.
 
For problem 1, 
% in the memory bank, the embedding of the GT frame can provide the most accurate template information, and the embedding of the latest frame changes over time, providing the most recent information for the segmentation of the query frame. However, the embedding of the latest frame lacks the use of historical information. 
% To this end, 
we propose a recurrent dynamic embedding (RDE) to provide a richer representation for VOS. As shown in Figure \ref{fig:fig1}(b), to generate and update RDE, we propose a  spatio-temporal aggregation module (SAM) to organize the cue of the historical information (previous RDE) and the embedding of the latest frame adaptively. 
SAM includes three parts: \textit{extracting}, \textit{enhancing} and \textit{squeezing}. The \textit{extracting} part is responsible for organizing the spatio-temporal relationship between previous RDE and the embedding of the latest frame. Then, the \textit{enhancing} part reinforces the spatio-temporal relationship and the \textit{squeezing} part aggregates and compresses the spatio-temporal information. We refer to the memory bank maintained by SAM as the SAM pattern memory bank. 

One potential risk of the SAM pattern memory bank is the recurrent update of RDE may cause error accumulation. However, we have no GT for training the generated RDE directly. To tackle this problem, we propose to employ auxiliary supervision for the distribution of RDE. In the training process, we additionally build a STM pattern memory bank (see Figure \ref{fig:fig1}(a)) to obtain the uncompressed information and its read results, which are used to estimate the distribution for RDE. Thus we design an unbiased guidance loss to control the approach degree of the two distributions. Relying on the unbiased guidance loss, the training of the network is more stable and has higher performance with no extra computation overhead for deployment.

% One potential risk of the SAM pattern memory bank is the recurrent use of SAM, because the update of RDE may cause error accumulation. However, we have no GT for training the generated RDE directly. Suppose the update process of the STM pattern memory bank is a good teacher, the estimated distribution read from the SAM pattern memory bank ought to approach the estimated distribution read from the STM pattern memory bank. Thus we design an unbiased guidance loss to control the approach degree of the two distributions. Relying on the unbiased guidance loss, the training of the network is more stable and has higher performance.

For problem 2, we design a novel self-correction strategy, which enforces the network to repair the embeddings of masks with different qualities in the memory bank. Specifically, we first simulate different perturbated masks and then constrain the embeddings encoded by perturbated masks to be close to the embedding encoded by the GT mask with a mask consistency loss. The mask consistency loss enforce the network to learn the self-correction ability for inaccurate masks in the embedding space during the training stage.

To investigate the effectiveness of the proposed methods, we conduct experiments on DAVIS 2017, DAVIS 2016 and YouTube-VOS 2019. The proposed method achieves state-of-the-art performance on DAVIS 2017 validation set (86.1\%  $\mathcal{J}$\&$\mathcal{F}$, 27 FPS), DAVIS 2017 test set (78.9\%  $\mathcal{J}$\&$\mathcal{F}$), DAVIS 2016  (91.6\% $\mathcal{J}$\&$\mathcal{F}$, 35 FPS) and superior performance on YouTube-VOS 2019 (83.3\% $\mathcal{J}$\&$\mathcal{F}$) without the multi-scale inference. Furthermore, we  demonstrate the effectiveness of our method in the synthetic long video. For the synthetic long video, $\mathcal{J}\&\mathcal{F}$ and FPS of our method are almost unchanged as the length of the synthetic long video increases.

Our contributions can be summarized as follows:
\vspace{-0.7em}
\begin{itemize}
    \item We propose an easy-to-extend recurrent dynamic embedding (RDE) to provide a richer representation for VOS compared with the embedding of the GT frame and the latest frame, which is maintained by the proposed spatio-temporal aggregation module (SAM).
    \vspace{-.5em}
    \item To avoid error accumulation owing to the recurrent usage of SAM, we propose an unbiased guidance loss during the training stage, which makes SAM  more robust in long videos. 
    \vspace{-.5em}
    \item Considering inaccurately predicted masks in the memory bank affect the segmentation performance due to the inaccurate network inference, we design a novel self-correction strategy, which enforces the network to learn the self-correction ability for inaccurate masks in the embedding space.
    \vspace{-.5em}
    \item Extensive experiments on several benchmarks and  the synthetic long video show the effectiveness and superiority of our method.
\end{itemize}
\vspace{-.5em}
% \end{itemize}

% %##################################################################################################
\begin{figure*}[!t]
\centering % 图片居中
\includegraphics[width=1\linewidth]{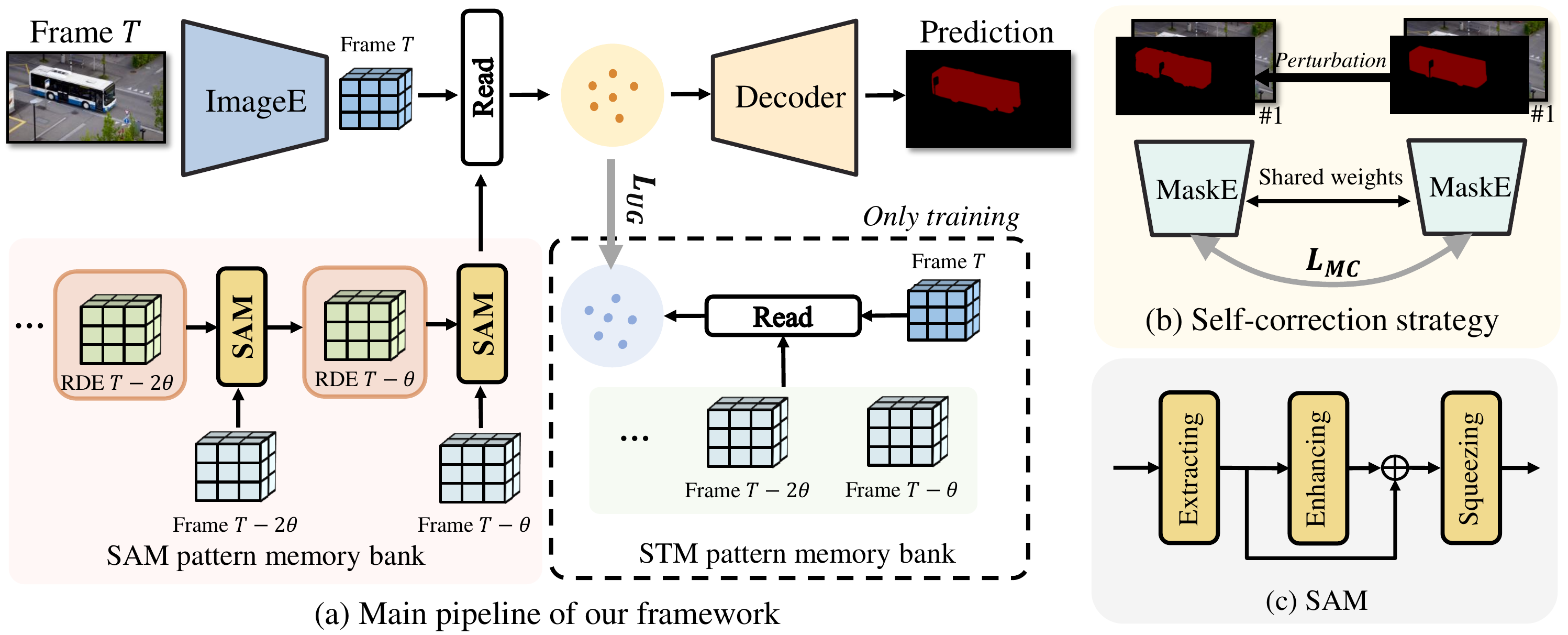}
\vspace{-1.2em}
\caption{
Illustrating the architectures: (a) The main pipeline of our framework. During the training stage, we maintain two individual memory banks which are updated in the STM pattern and our SAM pattern separately. During the inference, we only utilize our SAM pattern memory bank. $\theta$ denotes the sampling interval for the update of the memory bank. (b) Self-correction strategy. The proposed mask consistency loss $L_{MC}$ enforces the mask encoder to learn the self-correction ability for the inaccurate masks. (c) The structure of SAM, which organizes the historical information and the embedding of the latest frame adaptively. }
\vspace{-1em}
\label{fig:network}
\end{figure*}
% %##################################################################################################
% %##################################################################################################

\section{Related Work \label{sec:related}}
\subsection{Semi-supervised VOS}
Semi-supervised VOS mainly focuses on propagating the certain object mask of one frame. It can be roughly divided semi-supervised VOS into three categories: 1) Online fine-tuning based methods \cite{xiao2018monet, maninis2018video}, which usually learn general segmentation features and fine-tune the network to the target video during the test time. 2) Propagation based methods \cite{luiten2018premvos, chen2020state}, which refine the target segmentation mask in a temporal label propagation way. 3) Matching based methods \cite{oh2019video, seong2020kernelized, zhang2020transductive, cheng2021rethinking, mao2021joint} which encode some frames into embeddings and store those embeddings in the memory bank to segment the query frame. 
% %##################################################################################################
\subsection{Matching based VOS network}
STM~\cite{oh2019video} is a popular network in matching based methods, which constructs a continuously updated memory bank of historical frames. Compared with using limited frames (the GT frame \cite{hu2018videomatch} or the latest frame \cite{perazzi2017learning}), memory bank excavates the information of historical frames even more. Recently, matching based networks have received widespread attention. \cite{seong2020kernelized, lu2020video, cheng2021modular, hu2021learning} improve the readout operation of the memory bank, \cite{xie2021efficient} applies a local attention with the help of the optical flow, \cite{ge2021video} utilizes the global and instance embedding learning to address multi-objects VOS. Although these methods have achieved satisfactory performance, they ignore two key problems: 1) As the number of video frames increases, the hardware cannot afford the ever-increasing memory requirements. 2) Storing lots of information inevitably introduces lots of noise, which is not conducive to reading the most important information from the memory bank.

% %##################################################################################################

% %##################################################################################################
% \vspace{-0.4em}
\subsection{Efficient VOS network}
\vspace{-0.3em}
The methods of efficient VOS usually belong to propagation based methods or matching based methods. SAT \cite{chen2020state} is one of propagation based methods, which deals with each object as a tracklet and segments the object via two feedback loops. OSMN \cite{yang2018efficient} is one of the matching based methods, which adopt the GT frame and the latest frame to guide the segmentation of the query frame with two modulators. Recently, the most popular inference setting for VOS is to save the feature embedding of historical frames every 5 frames (STM pattern). Some methods \cite{li2020fast, liang2020video, wang2021swiftnet} try to use exponential moving average (EMA) to build a more efficient characterization to record the historical information. However, these methods only perform between the most similar embeddings due to the direct summation operation (see details in Sec. \ref{subsec:revisit}), which is a strong limitation. 
% %##################################################################################################

%##################################################################################################
% \vspace{-0.5em}
\section{Method}
%##################################################################################################
\subsection{Revisit Memory Bank Update with EMA}
\label{subsec:revisit}
In STM \cite{oh2019video}, the image and mask are encoded into two embedding spaces, named \textit{key} and \textit{value}. In addition to the \textit{key} and \textit{value} of the GT frame and the latest frame, previous EMA based methods build an independent embedding, $IE$. Take \textit{key} update as an example, let $\textbf{k}_t^{IE}(p)$ denotes the \textit{key} at time $t$ and $\textbf{k}^{Q}(q)$ denotes the \textit{key} of the  query frame  $Q$, where $p$ and $q$ are the coordinate of the spatial position. \cite{li2020fast, liang2020video} utilize EMA to update the historical embedding $\textbf{k}^{IE}_{t-\theta}(p)$ with the query embedding  $\textbf{k}^{Q}(q)$ by certain rules (see details in the supplementary material). The new embedding $\textbf{k}^{IE}_t(p)$ in the memory bank can be formulated as follows:
\begin{equation}
\textbf{k}^{IE}_t(p) = (1-\lambda)\textbf{k}^{Q}(q) + \lambda \textbf{k}^{IE}_{t-\theta}(p)
\label{eq:EMA}
\end{equation}
where $\lambda$ is a hyper-parameter to control the update strength and $\theta$ denotes the update interval. We argue that EMA based methods have a strong limitation, in which the two additional items in Eq. \ref{eq:EMA} must be similar in the parameter space because of the summation operation. Thus these methods \cite{li2020fast, liang2020video} index the most similar embeddings to update.  Our method associates the embeddings to update the extra embedding adaptively.
%##################################################################################################
\subsection{Framework Overview}
\paragraph{Encoders.}
The main pipeline of our framework is illustrated in Fig \ref{fig:network}(a). For a query frame of size $H \times W$, the image encoder $ImageE$ is responsible for extracting image features. We also adopt a mask encoder $MaskE$ to encode a certain frame and its mask to store into the memory bank. Both the encoders adopt ResNet-50 \cite{he2016deep} as the backbone and use two simple projection heads following STM \cite{oh2019video} to obtain two embeddings, \textit{key} $\textbf{k}\in \mathbb{R}^{C_k \times \frac{H}{16} \times \frac{W}{16}}$ and \textit{value} $\textbf{v} \in \mathbb{R}^{C_v \times \frac{H}{16} \times \frac{W}{16}}$. Here $C_k$ and $C_v$ are the numbers of the channel dimension ($C_k=64$, $C_v=512$ in our experiments). 

\vspace{-0.5em}
\paragraph{Memory Reading and Decoder.}
%During the training stage, we maintain two individual memory banks for the segmentation of the query frame, which is updated in the STM and SAM patterns separately. 
Following STCN \cite{cheng2021rethinking}, for the SAM pattern memory bank $m$ at time $t$,  we keep target-agnostic key $\textbf{k}^m_t$ and target-specific value $\textbf{v}^m_{t,i}$, where $i$ denotes the $i$-th object. For the similarity $\textbf{S}(p,q)$ between the key from the SAM pattern memory bank $\textbf{k}^m_t(p)$ and the key of the query frame $\textbf{k}^Q_t(q)$, we perform negative squared Euclidean distance, which can be formulated as 
\begin{equation}
    \label{eq:stm_similarity}
    \textbf{S}(p,q) = -||\textbf{k}^m_t(p)-\textbf{k}^Q_t(q)||^2_2
\end{equation}
where $p$ and $q$ are the coordinate of the spatial
position of $\textbf{k}^m_t(p)$ and $\textbf{k}^Q_t(q)$ separately. And the softmax operation is applied on the spatial dimension for similarity $S$ to obtain the  softmax-normalized affinity matrix $\textbf{W}$, $\textbf{W} = softmax(\textbf{S})$. Relying on $\textbf{W}$, the readout feature $\textbf{v}^{m\rightarrow Q}_{t,i}$ of the $i$-th object from the SAM memory bank can be obtained by the matrix multiplication $\odot$: 
\begin{equation}
    \label{eq:stm_reweight}
    \textbf{v}^{m\rightarrow Q}_{t,i} =  \textbf{W} \odot \textbf{v}^m_{t,i}.
\end{equation}
The readout feature $\textbf{v}^{m\rightarrow Q}_{t,i}$ concatenates with the value of the query frame to pass through the light-weight decoder described in \cite{cheng2021rethinking} to get the segmentation results $\Tilde{\textbf{y}}^{m}_{t, i}$ of the $i$-th object at frame $t$. Similar to the SAM pattern memory bank, we concatenate the readout feature $\textbf{v}^{M\rightarrow Q}_{t,i}$ from the STM pattern memory bank $M$ with the value of the query frame to obtain the segmentation results $\Tilde{\textbf{y}}^{M}_{t, i}$ of the $i$-th object at frame $t$.

\subsection{SAM Pattern Memory Bank}
The main challenge of keeping the size of the memory bank constant is how to select the most useful information. The STM pattern memory bank can store the historical information losslessly, but has ever-increasing size and inevitably introduces lots of noise. In our design, we build a SAM pattern memory bank to address the challenge. During the training stage, the STM and SAM pattern memory banks are maintained at the same time. During the inference, we only use the SAM pattern memory bank, which can keep the size of the memory bank constant. Specifically, the STM pattern memory bank $M$ includes $\{\textbf{k}^M_{t}, \textbf{v}^M_{t,i}\}$, while the SAM pattern memory bank $m$ includes $\{\textbf{k}^m_{t}, \textbf{v}^m_{t,i}\}$.  
\paragraph{Recurrent Dynamic Embedding.}
We find the embedding of the latest frame changes over time, providing more useful information for the segmentation of the query frame but lacking the use of historical information. We propose a recurrent dynamic embedding (RDE) in the memory bank to fuse the the cue of the historical information with the the embedding of the latest frame to provide a richer representation for VOS. We denote the RDE embedding at time $t$ as $\{\textbf{k}^{RDE}_{t}, \textbf{v}^{RDE}_{t,i}\} \in \{\textbf{k}^m_{t}, \textbf{v}^m_{t,i}\}$.  
\paragraph{Spatio-temporal Aggregation Module.} 
To generate and update RDE, we propose a spatio-temporal aggregation module (SAM), which exploits the cue of historical information.  SAM includes three parts: \textit{extracting}, \textit{enhancing} and \textit{squeezing} as shown in Figure \ref{fig:network}(c). The \textit{extracting} part is responsible for organizing the spatio-temporal relationship between the embedding of previous RDE $\{\textbf{k}^{RDE}_{t-\theta}, \textbf{v}^{RDE}_{t-\theta,i}\}$ ($\theta$ denotes the sampling interval) and the embeddings of the latest frame $\{\textbf{k}^L_{t}, \textbf{v}^L_{t,i}\}$. First, we concatenate previous RDE $\{\textbf{k}^{RDE}_{t-\theta}, \textbf{v}^{RDE}_{t-\theta,i}\}$ and the embedding of the latest frame $\{\textbf{k}^L_{t}, \textbf{v}^L_{t,i}\}$ to obtain the feature $\textbf{x}$. Take $\textbf{k}^{RDE}_t$ update as an example,  
\begin{equation}
    \textbf{x} = Cat(\textbf{k}^{RDE}_{t-\theta}, \textbf{k}^L_{t}), \textbf{x} \in \mathbb{R}^{C_k \times 2 \times \frac{H}{16} \times \frac{W}{16}}
\end{equation}
where $Cat$ denotes concatenation operation in the time dimension. Inspired by self-attention mechanism \cite{wang2018non}, in the \textit{extracting} part, we organize the spatio-temporal relationship between  previous RDE $\textbf{k}^{RDE}_{t-\theta}$ and the embedding of the latest frame $\textbf{k}^{L}_{t-\theta}$ to obtain the aggregation feature $\textbf{x}_{agg}$,
\begin{equation}
    \textbf{x}_{agg} = \frac{1}{C(\textbf{x})}\omega(\textbf{x})^\mathrm{T} \phi(\textbf{\textbf{x}}\downarrow) g(\textbf{x}\downarrow). 
\end{equation} 
$C(\textbf{x})$ is a normalization factor, which presents the total number of the spatial position of $\textbf{x}$. The function $\omega$, $\phi$ and $g$ are $1\times 1 \times 1$ convolution in our implementation. $\textbf{x} \downarrow$ denotes $\textbf{x}$ processed by the max-pooling operation (no down sampling on the time axis), which can decrease the computational complexity. 

Relying on the aggregation feature $\textbf{x}_{agg}$, in the \textit{enhancing} part, we enhance $\textbf{x}_{agg}$  in the form of residuals by atrous spatial pyramid pooling (ASPP) \cite{chen2017deeplab}. Finally, in the \textit{squeezing} part, we compress the enhanced feature by a simple $2\times 3\times 3$ convolution, which is denoted as  $Squeeze$ function. The formula can be expressed as
\begin{equation}
    \textbf{k}^{RDE}_t = Squeeze(\textbf{x}_{agg}+ASPP(\textbf{x}_{agg})).
\end{equation}
previous RDE and the embedding of the latest frame adaptively fuse and maintain the constant size for the memory bank, where the key mapping is $\mathbb{R}^{C_k \times 2 \times \frac{H}{16} \times \frac{W}{16}}$ $\rightarrow$ $\mathbb{R}^{C_k \times 1 \times \frac{H}{16} \times \frac{W}{16}}$. For the multiple objects, we concatenate the object dimension to the batch dimension like the implementation of STM \cite{oh2019video}. For the key and value of RDE, we maintain two different SAMs respectively.

% \paragraph{Memory Reading and Decoder.}
% After obtaining the RDE at time $t$, the memory reading of the SAM pattern memory bank is similar to the  memory reading of the STM pattern memory bank. The formula can be expressed as follows:
% \begin{equation}
%     \label{eq:sam_similarity}
%     \textbf{S}(p,q) = -||\textbf{k}^m_t(p)-\textbf{k}^L_t(q)||^2_2
% \end{equation}
% \begin{equation}
%     \label{eq:sam_reweight}
%     \textbf{v}^{m\rightarrow Q}_{t,i} =  \textbf{W} \odot \textbf{v}^m_{t,i}.
% \end{equation}
% Like the memory reading of the STM pattern memory bank, we concatenate the readout feature of the SAM pattern memory bank $\textbf{v}^{m\rightarrow Q}_{t,i}$ with  the value of the query frame to pass though the decoder to get the segmentation results $\Tilde{\textbf{y}}^{m}_{t, i}$ of $i$-th object at frame $t$. For the key and value of RDE, we maintain two different SAMs respectively.

% \vspace{-0.5em}
\paragraph{Unbiased Guidance Loss.}
One potential risk of the SAM pattern memory bank is the update of RDE may cause error accumulation, especially when it is used repeatedly. Another problem is that the key and value of RDE are generated separately by two different SAMs, the distribution of them is difficult to directly define. Suppose the update process of the STM pattern memory bank is a good teacher, the estimated distribution read from the SAM pattern memory bank ought to approach the estimated distribution read from the STM pattern memory bank. 
Thus, during the training stage, we maintain two individual memory banks for the segmentation of the query frame, which is updated in the STM and SAM patterns separately.
We propose an unbiased guidance loss $L_{UG}$, which controls the distribution of the readout feature from the SAM pattern memory bank $\textbf{v}^{m \rightarrow Q}_{t,i}$ to approach the distribution of the readout feature from the STM pattern memory bank $\textbf{v}^{M \rightarrow Q}_{t,i}$. The unbiased guidance loss $L_{UG}$ is computed as follows:
\begin{equation}
    % \small
    L_{UG} = \sum_i  KL (\textbf{v}^{M \rightarrow Q}_{t,i} || \textbf{v}^{m \rightarrow Q}_{t,i}). 
    \label{eq:ugloss}
\end{equation}
$KL$ function denotes Kullback–Leibler (KL) divergence, which is a non-symmetric measure of the difference between two distributions.

% \begin{figure}[t]
% \centering % 图片居中
% \includegraphics[width=.95\linewidth]{figure/inference.pdf}
% \vspace{-.3em}
% \caption{An illustration of the inference strategy with SAM. 	\raisebox{.46pt}{\textcircled{\raisebox{-.9pt} {t}}} denotes the key and value of the query frame at time $t$ encoded by the mask encoder $MaskE$. For brevity, we ignore the value of RDE to show the key update. SAM inputs previous RDE at time $t-\theta$ and the embeddings of the query frame at time $t$ to generate RDE at time $t$.}
% \label{fig:inference}
% \vspace{-1em}
% \end{figure}
% \vspace{-0.7em}

% \vspace{-0.2cm}
\paragraph{Self-correction Strategy.}
Considering the quality of the mask in the memory bank affects the segmentation of the query frame, we propose a mask consistency loss $L_{MC}$  to constrain the consistency of the embedding of masks of different qualities and the embedding of the GT mask.  
We first obtain the key $\textbf{k}_{1}$ and value $\textbf{v}_{1,i}$ of the first frame. And we perform perturbation transform such as the random dilation and eroding on the first frame to obtain the perturbated key $\ddot{\textbf{k}}_{1}$ and perturbated value $\ddot{\textbf{v}}_{1,i}$. The mask consistency loss $L_{MC}$ can be calculated by 
\begin{equation}
L_{MC} = KL (\textbf{k}_{1} || \ddot{\textbf{k}}_{1}) + \sum_i  KL (\textbf{v}_{1,i}|| \ddot{\textbf{v}}_{1,i})
\end{equation}
where $KL$ function denotes KL divergence.

% \vspace{-1cm}
\paragraph{Overall Loss Functions.}
During the training stage, we sample 5 frames. Inspired by the slowfast network \cite{feichtenhofer2019slowfast}, we utilize the SAM pattern memory bank to segment the third and fifth frames to handle different rate of videos. Besides, we utilize STM pattern memory bank to segment the second and fourth frames for the training stability.  We use bootstrapped cross entropy (BCE) following \cite{cheng2021modular} to supervise the final segmentation results, which is computed as follows:
\begin{equation}
\begin{split}
L_{Seg} = \frac{1}{2} (\sum_i \sum_{t=2,4} \underbrace{BCE(\Tilde{\textbf{y}}^{M}_{t,i}, \textbf{y}_{t,i})}_{STM  \; pattern \; item} +  \\
\sum_i \sum_{t=3,5} \underbrace{BCE(\Tilde{\textbf{y}}^{m}_{t,i}, \textbf{y}_{t,i})}_{SAM  \;  pattern \; item})
\label{eq:segloss}
\end{split}
\end{equation}
where $\Tilde{\textbf{y}}^{M}_{t,i}$ and $\Tilde{\textbf{y}}^{m}_{t,i}$ denote the segmentation results read from the STM pattern memory bank and the SAM pattern memory bank separately. $\textbf{y}_{t,i}$ denotes the GT mask of the $i$-th object at frame $t$. The overall loss function is computed as follows:
\begin{equation}
Loss = L_{Seg} +\mathbbm{1}[t=3,5] \mu L_{UG} + \gamma L_{MC}
\label{eq:loss}
\end{equation}
where $\mu$ and $\gamma$ are hyper-parameters to control the strength. We set $\mu=10$ and $\gamma=10$ in our experiments. $\mathbbm{1}[\cdot]$ is the indicator function.

\paragraph{Inference Strategy.}
As shown in Figure \ref{fig:network}(a), during the inference, we employ SAM recurrently to update RDE. Specifically, in a video of any length, SAM inputs previous RDE at time $t-\theta$ and the embeddings of the latest frame at time  $t$ to generate RDE at time $t$, where $\theta$ is the sampling interval. The new RDE is stored in the SAM pattern memory bank to assist the segmentation of the query frame and the old RDE is discarded.

% %##################################################################################################
% \begin{equation}
% z = ImageE(x)
% \label{eq:}
% \end{equation}

% $z = \{z_4, z_8, z_{16}\}$

% \begin{equation}
% \textbf{k}_m, \textbf{v}^i_m = Fuse(z_{16}, MaskE(x, \Tilde{\textbf{y}}^i))
% \label{eq:}
% \end{equation}

% \begin{equation}
% \textbf{v}^i_{m \rightarrow Q} = Retrieval(\textbf{k}_q | \textbf{k}_m, \textbf{v}^i_m)
% \label{eq:}
% \end{equation}

% \begin{equation}
% \Tilde{\textbf{y}}^i = Decoder(z | \textbf{v}^i_{m \rightarrow Q})
% \label{eq:}
% \end{equation}

% %##################################################################################################
% \subsection{Image Encoder}

% \begin{equation}
% \textbf{v}^i_{m \rightarrow Q} = Retrieval(\textbf{k}_q | \textbf{k}_m, \textbf{v}^i_m)
% \label{eq:}
% \end{equation}

% \begin{equation}
% \Tilde{\textbf{y}}^i = Decoder(z | \textbf{v}^i_{m \rightarrow Q})
% \label{eq:}
% \end{equation}

%##################################################################################################

%##################################################################################################

%##################################################################################################

\section{Experiments}
\subsection{Datasets and Metrics}
\paragraph{DAVIS.} DAVIS 2016 \cite{davis16} is a popular benchmark for  video single object segmentation, whose validation set includes  20 videos. DAVIS 2017 \cite{davis17} is a popular benchmark for video multiple objects segmentation, whose validation set and test set are 30 densely annotated videos. 
% Compared with DAVIS 2016, DAVIS 2017 is more challenging because of the introduction of multiple objects. 
\vspace{-0.3em}
\paragraph{YouTube-VOS.} YouTube-VOS 2019 \cite{youtube} is a large-scale benchmark for multi-object video segmentation, providing 3,471 videos for the training (65 categories) and 507 videos for the validation. There are additional 26 unseen categories in the validation set for evaluating the generalization.
\vspace{-0.3em}
\paragraph{Metrics.}
For the DAVIS datasets, we use the region similarity $\mathcal{J}$, the contour accuracy $\mathcal{F}$ and their average $\mathcal{J}\&\mathcal{F}$ to evaluate the segmentation results. For YouTube-VOS 2019, we follow the official evaluation server to report  $\mathcal{J}$ and $\mathcal{F}$ of the seen and unseen categories, and the average of them.

%##################################################################################################
\subsection{Implementation Details}
\paragraph{Training Stages.}
Following STCN \cite{cheng2021rethinking}, we first train the network equipped with the STM pattern memory bank on the static datasets \cite{wang2017learning,shi2015hierarchical,zeng2019towards,cheng2020cascadepsp,li2020fss} with 75k iterations and batch size of 64.  The static images are processed by synthetic deformations like STM \cite{oh2019video}. Secondly, we train the network quipped with the SAM and STM pattern memory banks on  BL30K \cite{chang2015shapenet,denninger2019blenderproc} proposed in \cite{cheng2021modular} with 500k iterations and batch size of 8. Finally, we fine-tune the network quipped with the SAM and STM pattern memory banks  on YouTube-VOS and DAVIS 2017 with 75k iterations and batch size of 16 (main stage). BatchNorm layers are frozen during the training stage following \cite{oh2019video}.
\vspace{-0.5em}
\paragraph{Training Details.} 
\label{sec:training details}
We adopt four 16 GB Tesla V100 GPUs to implement Pytorch. All networks are optimized by Adam optimizer \cite{kingma2014adam}. We pretrain the network on the static datasets and  BL30K  with the initial learning rate of 2e-5 and 1e-5. And we fine-tune the network on the main stage with the initial learning rate of 2e-5. The data augmentation is the same as STCN \cite{cheng2021rethinking}. Besides, we sample 3 frames in the first pre-training stage and 5 frames in other stages.
\vspace{-0.5em}
\paragraph{Inference Details.} 
\label{sec:Inference details}
During inference, we only use the SAM pattern memory bank. Specifically, in addition to maintaining our RDE by SAM, we sample the embedding of the latest frame and two repeated embedding of the GT frame. This setting is to keep a sampling balance between the accurate template information (GT frame) and dynamic information (latest frame or our RDE). We use top-k filters \cite{cheng2021modular} $k=40$ on all datasets. The sampling interval $\theta$ is set to 3 on all DAVIS datasets and 4 on YouTube-VOS 2019.  
% Other trivial hyper-parameters will be given in our released code. 

%##################################################################################################

\begin{table}[t]
% \small
\centering
\begin{tabular}{l c c c c c}
\hlineB{3}
Method & CC & {$\mathcal{J}$\&$\mathcal{F}$} & $\mathcal{J}$ & $\mathcal{F}$ & FPS\\ \midrule
% TVOS$^\dag$ ~\cite{zhang2020transductive}&$\times$ &72.3&69.9&74.7&\textbf{37}\\
STM$^\dag$~\cite{oh2019video}&$\times$&81.8&79.2&84.3&10.2\\
KMN$^\dag$ \cite{seong2020kernelized}&$\times$&82.8 &80.0 &85.6 & \textless 8.4\\
JOINT$^\dag$ \cite{mao2021joint} &$\times$&83.5 & 80.8& 86.2 &4.0 \\
LCM$^\dag$~\cite{hu2021learning}&$\times$&83.5 &80.5 &86.5 & \textless 8.5 \\
RMNet$^\dag$ \cite{xie2021efficient} &$\times$&83.5&81.0&86.0&\textless 11.9\\
MiVOS$^{\dag \ast}$  \cite{cheng2021modular} 
&$\times$& 84.5  & 81.7 & 87.4 & 11.2\\
HMMN$^\dag$ \cite{seong2021hierarchical}&$\times$&84.7 &81.9 &87.5 & \textless 10.0 \\
STCN$^{\dag \ast}$  \cite{cheng2021rethinking} &$\times$& \textbf{85.3}  & \textbf{82.0} & \textbf{88.6} & \textbf{20.2} \\ \hline
% SiamMask~\cite{wang2019fast}&$\surd$&56.4&64.3&58.5&35\\ 
GCNet~\cite{li2020fast}&$\surd$&71.4&69.3&73.5&\textless 25.0\\
% FEELVOS$^\dag$~\cite{voigtlaender2019feelvos}&$\surd$&71.5&69.1&74.0&2.2\\
% SAT~\cite{chen2020state}&$\surd$&72.3&68.6&76.0&39\\
Liang \textsl{et al.}~\cite{liang2020video}   &$\surd$& 74.6  & 73.0 & 76.1 & 4.0\\ 
G-FRTM$^\dag$ \cite{park2021learning} &$\surd$&76.4 & -&- & 18.2 \\
PReMVOS~\cite{luiten2018premvos} & $\surd$ & 77.8& 73.9& 81.7 & 0.01\\
SwiftNet$^\dag$~\cite{wang2021swiftnet}  &$\surd$& 81.1 & 78.3 & 83.9 &\textless 25.0\\ 
SST$^\dag$~\cite{duke2021sstvos} &$\surd $&82.5 & 79.9& 85.1 & -\\ 
Ge \textsl{et al.}$^\dag$~\cite{ge2021video}  &$\surd$&82.7 & 80.2& 85.3 & 6.7\\
\textbf{RDE-VOS}$^\dag$  &$\surd$&  84.2  & 80.8  & 87.5  & \textbf{27.0} \\
\textbf{RDE-VOS}$^{\dag \ast}$   &$\surd$& \textbf{86.1} & \textbf{82.1} & \textbf{90.0} & \textbf{27.0} \\
\hlineB{3}
\end{tabular}
\caption{Results on the DAVIS 2017 validation set. CC denotes constant cost during the inference. $^\dag$ indicates YouTube-VOS \cite{youtube} is added during the training stage. $^{\ast}$ denotes BL30K \cite{cheng2021modular} is added during the training stage.}
\label{tab:davis17}
\end{table}
\begin{table}[t]
	\centering
	\setlength{\tabcolsep}{1.8mm}{\begin{tabular}{lcccccc}
		\hlineB{3}
		Method
		 &CC& 600p & $\mathcal{J}\&\mathcal{F}$ & $\mathcal{J}$ & $\mathcal{F}$  \\
		\midrule
		STM$^\dag$~\cite{oh2019video} &$\times$ & $\surd$ & 72.2 & 69.3 & 75.2 \\
		KMN$^\dag$ ~\cite{seong2020kernelized} &$\times$ & $\surd$ & 77.2 & 74.1 & 80.3\\
		RMNet$^\dag$ ~\cite{xie2021efficient}&$\times$  &  $\times$ & 75.0 & 71.9 & 78.1  \\
		Ge \textsl{et al.}$^\dag$~\cite{ge2021video}&$\times$  &$\times$&75.2 & 72.0& 78.3 \\
% 		LCM$^\dag$~\cite{hu2021learning} &$\times$ & $\times$ & \textbf{78.1} & \textbf{74.4} & \textbf{81.8} \\
		STCN$^{\dag \ast}$  \cite{cheng2021rethinking} &$\times$ & $\times$ & 77.8 & 74.3 & 81.3\\ 
		MiVOS$^{\dag \ast}$  \cite{cheng2021modular}  &$\times$ & $\times$ & \textbf{78.6} & \textbf{74.9} & \textbf{82.2} \\\hline
		CFBI$^\dag$ ~\cite{yang2020collaborative} &$\surd$ & $\times$ & 74.8 & 71.1 & 78.5 \\
		Ge \textsl{et al.}$^\dag$~\cite{ge2021video} &$\surd$  &$\times$&75.2 & 72.0& 78.3\\
		CFBI+$^\dag$ ~\cite{yang2021collaborative} &$\surd$ & $\times$ & 75.6 & 71.6 & 79.6 \\
		\textbf{RDE-VOS}${^{\dag}}$ &$\surd$  & $\times$ & 77.4 & 73.6 & 81.2 \\
		\textbf{RDE-VOS}${^{\dag \ast}}$ &$\surd$  & $\times$ & \textbf{78.9} & \textbf{74.9} & \textbf{82.9}\\
		\hlineB{3}
	\end{tabular}}
	\caption{Results on the DAVIS 2017 test set. 600p denotes evaluating on 600p resolution.
% 	$^\dag$ indicates YouTube-VOS \cite{youtube} is added during the training stage. $^{\ast}$ denotes BL30K \cite{cheng2021modular} is added during the training stage. 
	}
	\vspace{-1em}
	\label{tab:davis17-test}
\end{table}

\subsection{Compare with the State-of-the-art Methods}
We denotes the memory bank of the constant size during the inference as \textbf{Constant Cost} (\textbf{CC}). As the video length increases during the inference, the CC methods can maintain a relatively stable speed and constant requirements of the memory.
For brevity, our \uline{R}ecurrent \uline{D}ynamic \uline{E}mbedding for  \uline{VOS} method is denoted as \textbf{RDE-VOS}. 
\vspace{-0.7em}
\paragraph{DAVIS.}
We compare the proposed method with previous state-of-the-art methods for VOS on the DAVIS 2017 validation set, DAVIS 2017 test set and DAVIS 2016 validation set.  On the DAVIS 2017 validation set, as shown in Table \ref{tab:davis17}, our method even outperforms STCN \cite{cheng2021rethinking} by 0.7\% for {$\mathcal{J}$\&$\mathcal{F}$} and runs about 35\% faster (27 \textit{vs} 20.2 FPS). Compared with SwiftNet \cite{wang2021swiftnet}, our method suppresses it by 5\%  for {$\mathcal{J}$\&$\mathcal{F}$} and has a slight advantage for the speed (+2 FPS). On the DAVIS 2017 test set, as shown in Table \ref{tab:davis17-test}, our method still has great advantages. On DAVIS 2016 validation set, as shown in Table \ref{tab:davis16}, our method outperforms CC method SwiftNet \cite{wang2021swiftnet} by 1.2\% for {$\mathcal{J}$\&$\mathcal{F}$} and runs about 40\% faster (35 \textit{vs} 25 FPS). Compared with STCN \cite{cheng2021rethinking}, our method is 30\% faster while {$\mathcal{J}$\&$\mathcal{F}$} is almost unchanged (-0.1\%). We also show the qualitative result of the validation set of DAVIS 2017 in Figure \ref{fig:results}. More qualitative results can be found in the supplementary material.

\begin{figure}[t]
\centering % 图片居中
\includegraphics[width=1\linewidth]{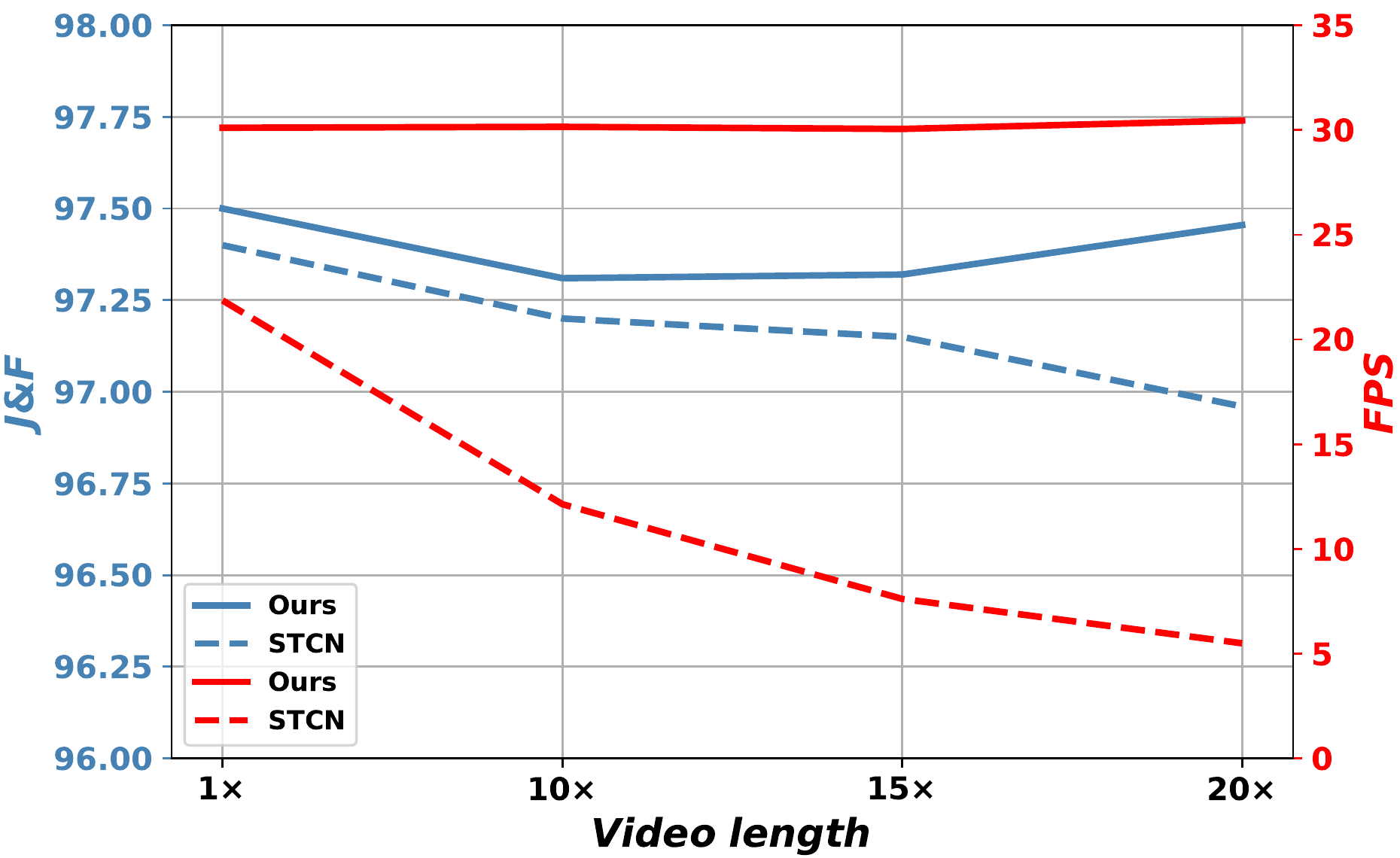}
\vspace{-1.5em}
\caption{
\textcolor[RGB]{70, 130, 180}{$\bm{\mathcal{J}\&\mathcal{F}}$} and  \textcolor[RGB]{255, 0, 0}{\textbf{\textit{FPS}}} of our method and STCN \cite{cheng2021rethinking} on the synthetic long video. Note different colored lines refer to different metrics. When the length of the synthetic long video is 1, 10, 15 and 20 times of the original, $\mathcal{J}\&\mathcal{F}$ and FPS of our method are almost unchanged. However, both $\mathcal{J}\&\mathcal{F}$ and FPS of STCN have an obvious reduction.}
\label{fig:long-video}
\vspace{-1em}
\end{figure}

\vspace{-0.5em}
\paragraph{YouTube-VOS.}
On a large-scale YouTube-VOS 2019 validation set, we compare our method with recent state-of-the-art methods in Table \ref{tab:youtube19}. Although our method does not surpass STCN on YouTube-VOS 2019 validation set, it still surpasses other state-of-the-art methods, regardless of whether BL30K is added.
\vspace{-1em}
\paragraph{Synthetic Long Video.}
Recently, the popular benchmarks include short video clips. For example, DAVIS 2017 only has 67 frames per video clip on average. However, many practical applications need to handle more frames. Compared with STCN \cite{cheng2021rethinking}, we demonstrate the effectiveness of our method in the scene where includes more frames. Take a DAVIS 2017 case ``cows" (the basic length is 104) as an example, exerting video forward and backward as a basic unit, we repeatedly sample multiple basic units to synthesize a long video. This synthesis method ensures each frame contains GT and the adjacent frames have smooth transitions. As shown in Figure~\ref{fig:long-video}, as the length of the synthetic long video increases, the performance and speed of our method is almost unaffected, while the performance and speed of STCN obviously decrease. Here we do not change any hyper-parameter compared with the setting on the DAVIS datasets. Besides, we utilize the official code of STCN and minimize the sampling interval to 60 frames under maximizing the usage of the GPU memory. All input data is stored in the CPU and inferred on one GPU.
 \begin{table}[t]
% \small
\centering
\begin{tabular}{l c c c c c}
\hlineB{3}
Method & CC & {$\mathcal{J}$\&$\mathcal{F}$} & $\mathcal{J}$ & $\mathcal{F}$ & FPS\\ \midrule
RMNet$^\dag$ \cite{xie2021efficient}&$\times$&88.8 &88.9 &88.7 & 11.9 \\
STM$^\dag$~\cite{oh2019video}&$\times$&89.3&88.7&89.9&6.3\\
KMN$^\dag$~\cite{seong2020kernelized}&$\times$&90.5 &89.5 &91.5 &8.4 \\
LCM$^\dag$~\cite{hu2021learning}&$\times$&90.7 &89.9 &91.4 &8.5 \\
HMMN$^\dag$ \cite{seong2021hierarchical} &$\times$&90.8 &89.6 &92.0 & 10.0 \\
MiVOS${^{\dag \ast}}$ \cite{cheng2021modular} &$\times$&91.0 &89.7 &92.4 & 16.9 \\
STCN${^{\dag \ast}}$  \cite{cheng2021rethinking} &$\times$& \textbf{91.7} &\textbf{90.4} &\textbf{93.0} & \textbf{26.9} \\
\midrule
GCNet~\cite{li2020fast}&$\surd$&86.6 &87.6 &85.7 & 25.0 \\
CFBI+$^\dag$~\cite{yang2021collaborative}&$\surd$&89.9 &88.7 &91.1 &5.9 \\
SwiftNet$^\dag$~\cite{wang2021swiftnet}&$\surd$&90.4 &\textbf{90.5} &90.3 & 25.0\\
\textbf{RDE-VOS}$^\dag$  &$\surd$& 91.1 & 89.7 & 92.5 & \textbf{35.0} \\
\textbf{RDE-VOS}$^{\dag \ast}$  &$\surd$& \textbf{91.6} & 90.0 & \textbf{93.2} & \textbf{35.0} \\
\hlineB{3}
\end{tabular}
\caption{Results on the DAVIS 2016 validation set. CC denotes constant cost during the inference.
% $^\dag$ indicates YouTube-VOS \cite{youtube} is added during the training stage. $^{\ast}$ denotes BL30K \cite{cheng2021modular} is added during the training stage.
}
\label{tab:davis16}
% \vspace{-1.5em}
\end{table}
% Youtube-vos 19
\begin{table}[!t]
% \small
\centering
\resizebox{1\columnwidth}{!}{
\begin{tabular}{l c c c c c c}
\hlineB{3}
Method & CC & Overall& $\mathcal{J}_{seen}$ & $\mathcal{F}_{seen}$ & $\mathcal{J}_{unseen}$ & $\mathcal{F}_{unseen}$ \\ \midrule
% STM~\cite{oh2019video}&$\times$&89.3&88.7&89.9&6.3\\
% CFBI+~\cite{yang2021collaborative}&$\times$&89.9 &88.7 &91.1 &5.9 \\
% KMN~\cite{seong2020kernelized}&$\times$&90.5 &89.5 &91.5 &8.4 \\
% LCM~\cite{hu2021learning}&$\times$&90.7 &89.9 &91.4 &8.5 \\
% RMNet$^\dag$~\cite{xie2021efficient}&$\times$&88.8 &88.9 &88.7 & 11.9 \\
% {\color{red} STCN \cite{cheng2021rethinking} }&$\times$& 91.7 &90.4 &93.0 & 26.9 \\
% \midrule
% GCNet~\cite{li2020fast}&$\surd$&86.6 &87.6 &85.7 & 25 \\
% SwiftNet~\cite{wang2021swiftnet}&$\surd$&90.4 &90.5 &90.3 & 25\\
STM$^\dag$ \cite{oh2019video} &$\times$ & 79.2 &79.6 & 83.6 & 73.0 & 80.6 \\
% HMMN$^\dag$ \cite{seong2021hierarchical} &$\times$ &82.5 &81.7 & 86.1 & 77.3 & 85.0 \\
% JOINT \cite{mao2021joint}  &$\times$&82.8 &80.8 & 86.1 &79.0 & 86.6 \\
MiVOS$^{\dag \ast}$ \cite{cheng2021modular} &$\times$ &82.4 &80.6 & 84.7 & 78.2 & 85.9 \\
STCN$^{\dag \ast}$ \cite{cheng2021rethinking} &$\times$&\textbf{84.2} &\textbf{82.6} & \textbf{87.0} &\textbf{79.4} & \textbf{87.7} \\ \hline
CFBI$^\dag$ \cite{yang2020collaborative} &$\surd$ &81.0 &80.6 & 85.1 & 75.2 & 83.0 \\
SST$^\dag$ \cite{duke2021sstvos}  &$\surd$& 81.8 &  80.9 & - & 76.6 & - \\  
\textbf{RDE-VOS $^{\dag}$} &$\surd$& 81.9  & 81.1 & 85.5   & 76.2  & 84.8 \\
\textbf{RDE-VOS $^{\dag \ast}$} &$\surd$& \textbf{83.3} & \textbf{81.9} & \textbf{86.3} & \textbf{78.0} & \textbf{86.9} \\
\hlineB{3}
% \vspace{-1em}
\end{tabular}}
\caption{Results on the YouTube-VOS 2019 validation set. 
% $^\dag$ indicates YouTube-VOS \cite{youtube} is added during the training stage. $^{\ast}$ denotes BL30K \cite{cheng2021modular} is added during the training stage.
}
\label{tab:youtube19}
\vspace{-1.5em}
\end{table}
\vspace{-1em}
\paragraph{Inference Time.}
We evaluate the inference time on one Tesla V100 GPU with full floating point precision. On the validation set of DAVIS 2017 and DAVIS 2016, as shown in Table \ref{tab:davis17} and \ref{tab:davis16}, our method has a great advantage in speed compared with STCN (27 \textit{vs} 20.2 FPS on DAVIS 2017 and 35 \textit{vs} 26.9 FPS on DAVIS 2016). 
%##################################################################################################

%####################
%##################################################################################################
\begin{figure*}[t]
% \subfigure{
% \begin{minipage}[t]{1\linewidth}
% \centering
% \includegraphics[width=1\linewidth]{figure/results1.pdf}
% %\caption{fig2}
% \end{minipage}
% }%
% \vspace{-1em}
% \subfigure{
% \begin{minipage}[t]{1\linewidth}
\centering
\includegraphics[width=1\linewidth]{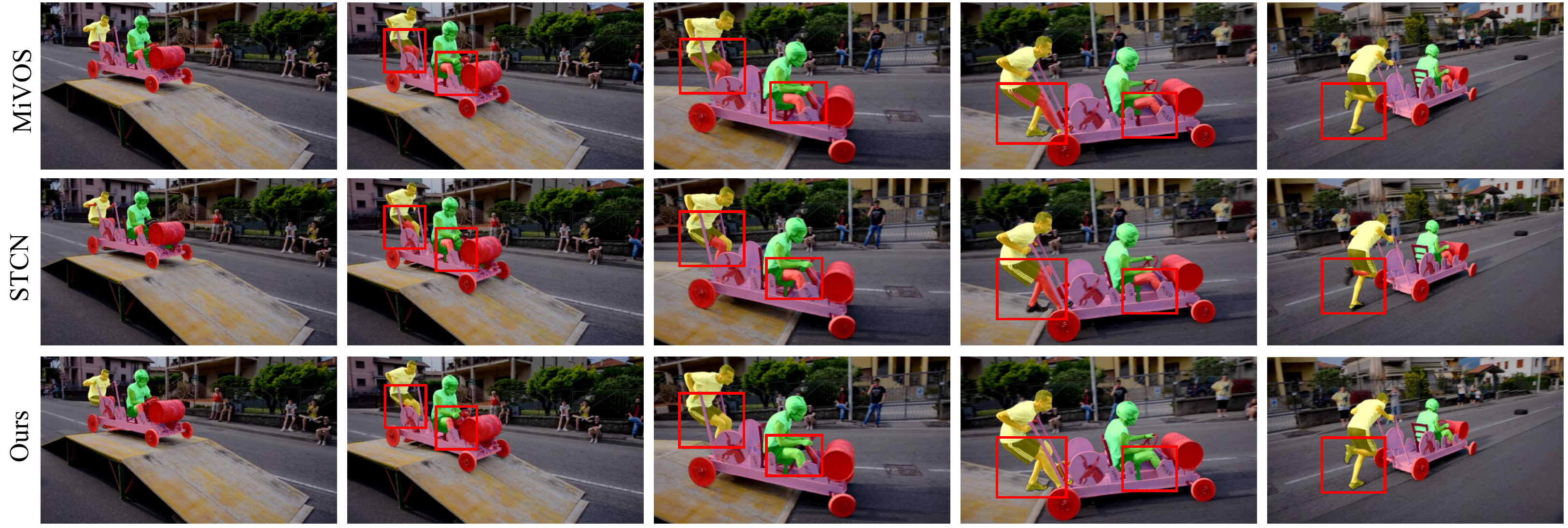}
%\caption{fig2}
% \end{minipage}
% }%
\caption{Qualitative results on the DAVIS 2017 validation set. We compare MiVOS \cite{cheng2021modular} and STCN \cite{cheng2021rethinking} under the challenging scale and deformation case, and our method has a notable improvement.}
\label{fig:results}
\vspace{-1.3em}
\end{figure*}
\subsection{Ablation Study}
\paragraph{Dataset Setting.}
We compare the results whether to adopt BL30K \cite{cheng2021modular} in Table \ref{tab:davis17}, \ref{tab:davis17-test}, \ref{tab:davis16} and \ref{tab:youtube19}. Without the BL30K pre-training, our method has superior performance on all datasets with a higher speed compared with other state-of-the-art methods. After adding the BL30K pre-training, our method has a stable improvement on all datasets.

% \begin{table}[!t]
% 	\centering
% 	\begin{tabular}{lccc}
% 	\hlineB{3}
%     Variants &$\mathcal{J}\&\mathcal{F}$&$\mathcal{J}$&$\mathcal{F}$ \\ \hline
%     RDE & 81.8 & 78.0 & 85.7 \\\hline
%     First frame & 71.6 & 67.8 & 75.4\\ 
%     First frame \& RDE & 85.3 & 81.6 & 89.0 \\\hline
%     Latest frame & 80.4 & 76.9 & 83.8\\
%     Latest frame \& RDE & 82.2 & 78.4& 86.0\\\hline
%     First frame \& latest frame & 84.6  & 81.0 & 88.2 \\
%     1F \& L \& RDE  & 85.4 & 81.6& 89.2 \\ \hline
%     First frame $\times$2  \& latest frame & 85.1  & 81.5 & 88.7 \\
%     First frame \& latest frame $\times$2 & 84.0  & 80.4 & 87.6\\
%     2F \& L \& RDE & \textbf{86.1} & \textbf{82.1} & \textbf{90.0} \\ 
%     \hlineB{3}
% 	\end{tabular}
% 	\caption{Ablation study of inference strategies on  DAVIS 2017 validation set. F \& L \& RDE denote first frame, latest frame and RDE. 2F denotes we sample the embeddings of the GT frame twice in order to balance the accurate template information and dynamic information, which is used in all experiments unless otherwise specified. }
% 	\label{tab:ablation-inference}
% \end{table}
% \vspace{-2em}
\paragraph{Inference Setting.}
\label{sec:ablation-inference}
Table \ref{tab:ablation-inference} shows different inference strategies adopting the memory bank. Compared with only using the embedding from the first frame or the latest frame, only using our RDE has the best performance of 81.8\% for {$\mathcal{J}$\&$\mathcal{F}$}. Besides, based on using the embeddings from the first frame, the latest frame, and both of them, adding our RDE can further improve  {$\mathcal{J}$\&$\mathcal{F}$} by 13.7\%, 1.8\% and 0.8\% separately.  Based on using RDE and the embedding of both the first frame and the latest frame, we explore the sampling balance of the accurate template information (GT frame) and dynamic information (latest frame or our RDE).  We find additionally sample the embedding of the GT frame to keep the sampling balance between the two types of information can further improve  {$\mathcal{J}$\&$\mathcal{F}$} by 0.7\%. We use this strategy in all experiments unless otherwise specified.
% The other hyper-parameters can be found in the supplementary material, such as sampling interval $\theta$.
We also show the ablation of different sampling intervals $\theta$, where the sampling interval of 3 provides the best result.
\vspace{-0.5em}
\paragraph{Loss Function Setting.}
In Table \ref{tab:ablation-loss}, we perform ablation of different loss functions without the BL30K \cite{cheng2021modular} pre-training. Both our proposed $L_{MC}$ and $L_{UG}$  can improve the performance to different degrees, and their combination can maximize the performance (+1.7\% {$\mathcal{J}$\&$\mathcal{F}$}). Besides, although we do not use the STM pattern memory bank during the inference, we find supervising the segmentation results of the STM pattern item  in Eq. \ref{eq:segloss} can assist the training of the SAM pattern (+1.2\%  {$\mathcal{J}$\&$\mathcal{F}$}). 

% \paragraph{Analysis of RDE.}
% In Figure \ref{fig:pca}, we show the memory bank embeddings of STCN and our method. The embeddings are reduced to 3 channels by PCA \cite{PCA} for visualization. Although the STM pattern memory bank stores the historical information losslessly, it brings lots of noise. The key and value of RDE are similar to the embeddings of latest frame in a different representation. The difference originates from the introducing of historical information (previous RDE). 
\begin{table}[!t]
	\centering
	\begin{tabular}{lccc}
	\hlineB{3}
    Variants &$\mathcal{J}\&\mathcal{F}$&$\mathcal{J}$&$\mathcal{F}$ \\ \hline
    \multicolumn{4}{c}{Strategy permutation} \\\hline
    RDE & 81.8 & 78.0 & 85.7 \\\hline
    First frame & 71.6 & 67.8 & 75.4\\ 
    First frame \& RDE & 85.3 & 81.6 & 89.0 \\\hline
    Latest frame & 80.4 & 76.9 & 83.8\\
    Latest frame \& RDE & 82.2 & 78.4& 86.0\\\hline
    First frame \& latest frame & 84.6  & 81.0 & 88.2 \\
    F \& L \& RDE  & 85.4 & 81.6& 89.2 \\ \hline
    First frame $\times$2  \& latest frame & 85.1  & 81.5 & 88.7 \\
    First frame \& latest frame $\times$2 & 84.0  & 80.4 & 87.6\\
    2F \& L \& RDE & \textbf{86.1} & \textbf{82.1} & \textbf{90.0} \\ \hline
    \multicolumn{4}{c}{Sampling interval $\theta$} \\\hline
    2F \& L \& RDE ($\theta=$ 2) & 85.1 & 81.4 & 88.9 \\ 
    2F \& L \& RDE ($\theta=$ 3) & \textbf{86.1} & \textbf{82.1} & \textbf{90.0} \\
    2F \& L \& RDE ($\theta=$ 4) & 85.1&81.5 &88.8\\
    2F \& L \& RDE ($\theta=$ 5) & 84.2&80.5 &87.9 \\
    \hlineB{3}
	\end{tabular}
	\caption{Ablation of inference strategies on DAVIS 2017 validation set. F \& L \& RDE represents first frame, latest frame and RDE. 2F represents we sample the embeddings of the GT frame twice in order to keep balance of the accurate template information and dynamic information, which is used in all experiments unless otherwise specified. }
	\label{tab:ablation-inference}
	\vspace{-1.3em}
\end{table}
\begin{table}[t]
	\centering
	\resizebox{1\columnwidth}{!}{
	\begin{tabular}{ccccc}
	\hlineB{3}
    \multicolumn{1}{c|}{} &Ablation Settings &  $\mathcal{J}\&\mathcal{F}$   &$\mathcal{J}$&$\mathcal{F}$\\ \hline
% 	  \multicolumn{1}{c|}{\multirow{2}{*}{Architecture}}  &SAM w/o extracting part &  &  &  \\ 
	\multicolumn{1}{c|}{\multirow{3}{*}{Loss}} & w/o $L_{MC}$ &83.7 & 80.5 & 86.9    \\
	 \multicolumn{1}{c|}{}  &w/o $L_{UG}$ & 82.9 & 79.5 & 86.4 \\ 
	 \multicolumn{1}{c|}{}  &w/o $L_{MC}$ \& $L_{UG}$ & 82.5 & 79.1 & 86.0 \\ 
	 \multicolumn{1}{c|}{}  & $L_{Seg}$ w/o STM pattern item  & 83.0 & 79.4 & 86.6 \\\hline
%   \multicolumn{1}{c|}{Architecture}  &SAM w/o enhancing part & 82.9 & 79.8 & 86.0  \\\hline
	 \multicolumn{1}{c|}{}  &Full &  \textbf{84.2} & \textbf{80.8}  & \textbf{87.5}\\
    % $L_{MC}$ &$L_{UG}$ &$\mathcal{J}$&$\mathcal{F}$& $\mathcal{J}\&\mathcal{F}$  \\ \hline
    % $\times$ & $\times$ & & & \\ 
    % $\surd$ & $\times$ & & & \\
    % $\times$  &$\surd$ & & & \\
    % $\surd$ &$\surd$ & & & \\
    \hlineB{3}
	\end{tabular}}
	\vspace{-0.5em}
	\caption{Ablation of different loss functions without the BL30K \cite{cheng2021modular} pre-training.}
	\label{tab:ablation-loss}
	\vspace{-1em}
\end{table}

\subsection{Limitations.}
During the inference, we set the sampling interval of the RDE update to $\theta$. This simple setting is easy to plug in other matching based VOS methods. We fix the sampling interval $\theta=3$ on the DAVIS datasets and achieve the new state-of-the-art performance. We increase the sampling interval on YouTube-VOS by 1 to fit the motion pattern on YouTube-VOS. 
% Note under this setting, the speed and performance of our method increase compared with the sampling interval 3. 
A better solution for future work is to use a learnable discriminator \cite{goodfellow2020generative} or gate mechanism \cite{yu2019free} to adaptively control the update interval of SAM, which can handle different scenes better.

\section{Conclusion}
In this paper, we explore how to build and update a memory bank of the constant size to maximize the segmentation performance of the query frame. The key insight is we propose a recurrent dynamic embedding (RDE) to provide a richer representation for VOS compared with the embeddings of the GT frame and the latest frame. To generate and update RDE, we propose a novel spatio-temporal aggregation module (SAM), which organizes the cue of the historical information and the embedding of the latest frame adaptively. To avoid error accumulation owing to the recurrent usage of SAM, we propose an unbiased guidance loss during the training stage, which makes SAM more robust in long videos. Besides, we design a novel self-correction strategy so that the network can encode and self-repair the embeddings of masks with different qualities. 
% We hope the RDE mechanism can help future matching based VOS works to further explore the potential of the memory bank of the constant size.
\paragraph{Acknowledgement.}
We acknowledge funding from National Key R\&D Program of China under Grant 2017YFA0700800, National Natural Science Foundation of China under Grants 61931014, 62131003 and 62021001, and Fundamental Research Funds for the Central Universities under No. WK3490000006.

% {\small
% \normalem
% \bibliographystyle{ieee}
% \bibliography{egbib}}%%我们的例子应该是\bibliography{cited}

% %%%%%%%%% REFERENCES
\small
\normalem
\bibliographystyle{ieee_fullname}
\bibliography{egbib}

\end{document}